%% file: iclr2027_conference.tex
\documentclass{article} 
\usepackage{iclr2027_conference,times}

\input{math_commands.tex}

\usepackage{booktabs}
\usepackage{array}
\usepackage{graphicx}  

\usepackage{caption}
\usepackage{amsmath,amsfonts,bm}
\usepackage{amsthm}      
\usepackage{amssymb}     

\usepackage{mathtools}   
\usepackage{algorithm}
\usepackage[algo2e, linesnumbered, ruled, vlined]{algorithm2e}
\usepackage{algpseudocode}
\SetKwInOut{Input}{Input}

\usepackage{wrapfig}             
\usepackage{lipsum} 

\theoremstyle{plain}

\newcommand{\mc}[1]{\mathcal{#1}}  
\newcommand{\tsc}[1]{\textsc{#1}}  
\newcommand{\ttt}[1]{\texttt{#1}}  

\usepackage{hyperref}       
\hypersetup{
    colorlinks = true,
    citecolor = LGRed,  
    linkcolor  = [HTML]{0668E1},  
    filecolor  = [HTML]{00008B},  
    urlcolor   = [HTML]{00008B}   
}

\usepackage{twemojis} 
\usepackage{fontawesome5}

\usepackage{url}
\usepackage{tcolorbox}
\tcbuselibrary{breakable,skins}  

\usepackage{multirow}
\usepackage[table]{xcolor}
\definecolor{mybrown}{HTML}{EA580C}
\definecolor{mygreen}{HTML}{008744}
\definecolor{myorange}{HTML}{ffa700}
\definecolor{myred}{HTML}{d62d20}
\usepackage{subcaption}

\providecolor{LGRedDark}{HTML}{7C0027}   
\providecolor{LGRed}{HTML}{A50034}       
\providecolor{LGRedLight}{HTML}{C04D71}  

\usepackage{makecell}
\usepackage{xcolor} 
\usepackage{subcaption} 

\usepackage{natbib}

\title{Continually Post-training Small Models\\via Synthetic Self-distillation}

\author{%
  Woosung Koh\textsuperscript{\scalebox{0.85}{$\spadesuit$}}*,
  Hyunsoo Lee\textsuperscript{$\Diamond$},
  Kyungjae Lee\textsuperscript{$\Diamond$},
  Haeju Park\textsuperscript{$\Diamond$},
  Dahyun Lee\textsuperscript{$\Diamond$},\\
  \bfseries Dasol Hwang\textsuperscript{$\Diamond$},
  Moontae Lee\textsuperscript{$\Diamond$}$^\dag$\\[6pt]
  \textsuperscript{\scalebox{0.85}{$\spadesuit$}}KAIST AI \quad \textsuperscript{$\Diamond$}LG AI Research\\[5pt]
  {\scriptsize*Work done during an internship at LG AI Research\quad%
   {$^\dag$}Corresponding author}
}

\ifdefined\ARXIVVERSION
  \iclrfinalcopy                        
  \renewcommand{\headrulewidth}{0pt}    
\fi

\iclrfinalcopy 
\begin{document}

\newcommand{\paperabstract}{%
Continual post-training faces a dual challenge of acquisition and retention: models must learn new capabilities while preserving prior knowledge. While Self-Distillation Fine-Tuning (SDFT) mitigates this using self-supervision with in-context demonstrations, it relies on strong in-context learning and instruction-following abilities. Consequently, SDFT fails on smaller models critical for low-latency, on-device use. To enable continual learning at smaller scales, we first identify the root cause of failure. We show that poor acquisition and forgetting correlate linearly with the post-trained model's distribution shift, revealing the necessity of in-distribution learning. We thus introduce Synthetic Self-Distillation (\tsc{SynthSelf}), imitating SDFT's in-distribution rewrite by outsourcing the task to a minimally larger, same-family aligner. Coupled with off-policy supervision and rejection sampling, this provides the high-quality supervision small models require without breaking their original distribution. Evaluations spanning \textbf{4} acquisition datasets (including a mixture) and \textbf{14} retention benchmarks show \tsc{SynthSelf} consistently outperforms \textbf{6} methods (including off- and on-policy distillation) across \textbf{3} small model scales. Extensive further analyses confirm \tsc{SynthSelf} paves the way toward effective small-model continual learning.
}

\ifdefined\ARXIVVERSION
  \input{arxiv_cover.tex} 
\else
  \maketitle
  \begin{abstract}
  \paperabstract
  \end{abstract}
\fi

\section{Introduction}

Continual post-training of language models (LMs) is a common setting where practitioners aim to inject new capabilities into already-trained models for scenarios such as customizing for specific use-cases and enabling human-like lifelong learning \citep{shi2024continual,wang-etal-2024-inscl, zheng2025spurious, 10.1145/3716629,srijith-etal-2025-continual, chen2026staticmodelsevolvingframework}. Yet, this process is notoriously challenging on both the acquisition (injecting new abilities) and retention (maintaining existing capabilities) axes \citep{ harmon2026mapping, 11151751, chen2026retaining}. The failure to retain existing capabilities, widely known as catastrophic forgetting \citep{MCCLOSKEY1989109, french1999catastrophic, doi:10.1073/pnas.1611835114}, is particularly well-documented in settings where a supervised fine-tuning (SFT) dataset is used for teacher forcing \citep{kalajdzievski2024scalinglawsforgettingfinetuning, 11151751, lai2026reinforcementfinetuningnaturallymitigates, chen2026retaining, shenfeld2026rls}.

A recent work showed that self-distillation fine-tuning (SDFT) is an effective way to inject SFT demonstrations in an on-policy manner \citep{shenfeld2026selfdistillation}, outperforming recent off-policy continual fine-tuning methods such as re-invoke \citep{lu2025onpolicydistillation} and Dynamic Fine-tuning (DFT; \citet{wu2026on}) on both acquisition and retention axes. SDFT leverages the in-context learning (ICL) and instruction-following (IF) abilities of post-trained models to provide demonstrations in-context to the self-teacher model, and applies on-policy distillation (OPD; \citet{pmlr-v15-ross11a}, \citet{agarwal2024onpolicy}) on itself, supervising on an implicit self-rewrite of the demonstrations.

However, as models struggle with ICL and IF at smaller scales, SDFT remains infeasible in this regime (see Fig.~\ref{fig:sdft-scale}). We aim to address this limitation, as small models have become critical for applications such as on-device deployment, latency-specialized models, draft models, verifiers, and large-small collaborative methods \citep{cobbe2021trainingverifierssolvemath, chen2023acceleratinglargelanguagemodel, abdin2024phi-, pmlr-v235-liu24ce, lu2025smalllanguagemodelssurvey, belcak2025small, cai2026think}.

\paragraph{Contribution.} To better understand how to leverage the key principles of SDFT, we first examine why it succeeds in continual post-training (\textbf{\S\ \ref{sec:dist_cont}}). We reveal that the distribution shift of the post-trained base model linearly correlates with poor acquisition and forgetting (\textbf{Fig.~\ref{fig:shift}}), indicating that staying in-distribution is crucial for continual learning. While the primary advantage of SDFT over off-policy methods is its ability to staying in distribution, evaluating its self-generated offline dataset reveals that SDFT's self-supervision quality is inadequate at smaller model scales (\textbf{Fig.~\ref{fig:sdft-selfrewrite}}).

\begin{figure*}[bt]
\centering
\begin{subfigure}[b]{0.49\textwidth}
  \centering
  \includegraphics[width=\linewidth]{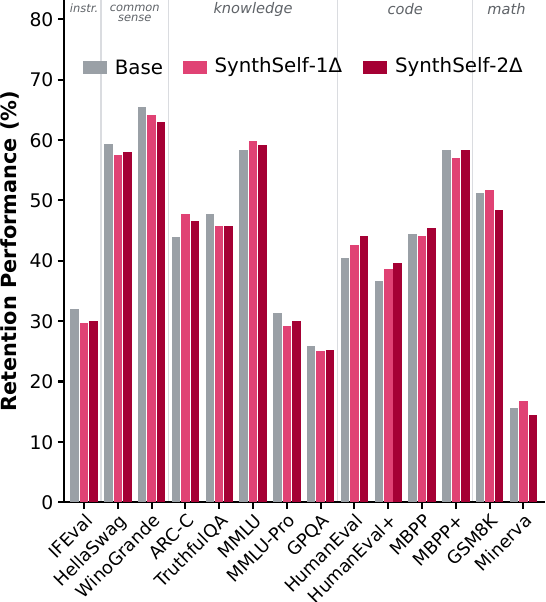}
  \subcaption{Averaged across 3 base models, continual post-training via \tsc{SynthSelf} on the medical domain results in near-zero forgetting across all 14 benchmarks.}
  \label{fig:retention}
\end{subfigure}
\hfill
\begin{subfigure}[b]{0.49\textwidth}
  \centering
  \includegraphics[width=\linewidth]{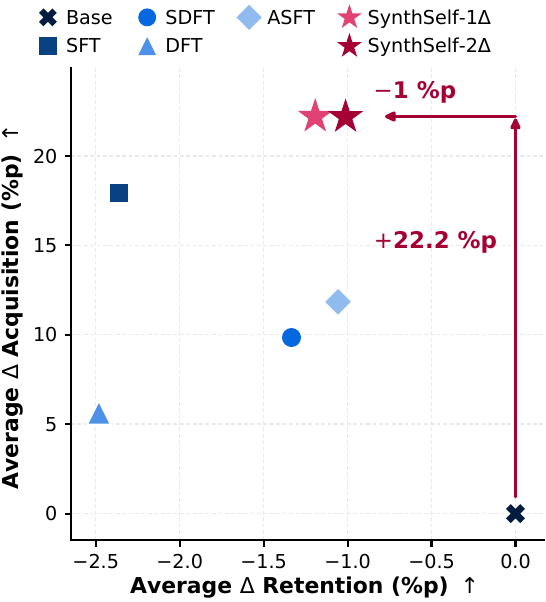}
\subcaption{Across 3 base models, 4 acquisition benchmarks, and 14 retention tasks, \tsc{SynthSelf} maximizes performance gains with minimal forgetting.}
  \label{fig:pareto}
\end{subfigure}
\caption{\textbf{Overview of \tsc{SynthSelf} performance across three model scales (\ttt{0.5}, \ttt{1.5}, and \ttt{3B}).}}
\vspace{-1.5em}
\label{fig:headline}
\end{figure*}

In response, we propose Synthetic Self-Distillation (\tsc{SynthSelf}), which \textit{imitates} self-distillation's implicit in-distribution rewrite by outsourcing the distribution-aligned rewrite to a larger model (\textbf{\S\ \ref{sec:synthself}}). Our method distinctly differs from SDFT in the following ways: \textbf{(i)} choosing off-policy over on-policy supervision allows us to apply rejection sampling, which in turn improves supervision quality for small models that are sensitive to data quality; and \textbf{(ii)} using minimally larger, same-family aligners allows us to simultaneously \textbf{(ii.a.)} provide high-quality supervision (\textbf{Tab.~\ref{tab:aligner-family}}), \textbf{(ii.b.)} remain in-distribution relative to the base model (\textbf{Fig.~\ref{fig:teacher_dist}}), and \textbf{(ii.c.)} maintain low computational costs. We empirically validate \tsc{SynthSelf} (\textbf{\S\ \ref{sec:exp}}), demonstrating that:

\begin{enumerate}

\item \textsc{SynthSelf} achieves the best acquisition across \textbf{3} domains (science, tool-use,
and medicine) and a 4-subject science mixture while preserving prior capabilities on
\textbf{14} benchmarks, consistently across \textbf{3} base model sizes (\texttt{0.5}, \texttt{1.5},
\texttt{3B}), and against \textbf{4} relevant baselines (\textbf{Fig.~\ref{fig:headline}, Tab.~\ref{tab:main}}).

\item Controlling for the confounding factor of teacher model size, we demonstrate that
\textsc{SynthSelf} outperforms both on- and off-policy distillation in acquisition by
\textbf{+5.7\%p} to \textbf{+16.9\%p}, averaged over all four settings and three base scales
(\textbf{Tab.~\ref{tab:main}}, \textbf{Fig.~\ref{fig:ablation}}).

\item We show that utilizing a smaller (weaker) in-distribution aligner is more effective than deploying a larger (stronger) out-of-distribution teacher, yielding a \textbf{69.9\%} reduction in alignment computational cost (FLOPs) (\textbf{Tab.~\ref{tab:aligner-family}}).

\item We validate that our improvement over SDFT is not due to prompt differences using prompt-matched and diverse prompts, achieving \textbf{+18.96\%p} and \textbf{+1.68\%p} gains in acquisition and retention over prompt-matched SDFT (\textbf{Fig.~\ref{fig:sdft_prompt}}).

\item Finally, our ablation studies confirm that combining rejection sampling with off-policy supervision is a meaningful component of our methodology (\textbf{Fig.~\ref{fig:ablation}}).
\end{enumerate}

\section{Method}

\subsection{Problem Setting: Continual Fine-tuning}

\paragraph{Continual fine-tuning.} Given a post-trained policy $\pi$, we continually inject new knowledge from a dataset $\mc{D}_\text{train} = \{(x_i, y_i)\}_{i=1}^N$ by further training $\pi$. In doing so, we aim to maximize \textbf{(i)} acquisition: performance on a same-domain test set $\mc{D}_\text{test}$, while maintaining \textbf{(ii)} retention: performance on existing capabilities, evaluated via $\mc{D}_\text{ret}$.

\paragraph{Self-distillation fine-tuning.}
Given a demonstration pair $(x,y)$ and an SDFT prompt $P^{\text{SDFT}}$ that places the demonstration in-context, SDFT trains the policy $\pi_\theta$ parameterized by $\theta$ on-policy toward the frozen model's self-rewrite of that demonstration,
\begin{equation}
\mathcal{L}(\theta)\;=\;D_{\mathrm{KL}}\big(\pi_\theta(\cdot\mid x)\,\big\|\,
\pi\!\left(\cdot \mid P^{\text{SDFT}}(x, y)\right)\big)
\;=\;\mathbb{E}_{y'\sim\pi_\theta(\cdot\mid x)}
\left[\log\frac{\pi_\theta(y'\mid x)}{\pi\!\left(y' \mid P^{\text{SDFT}}(x, y)\right)}\right],
\label{eq:sdft}
\end{equation}
where $y'$ denotes a completion sampled from the current policy. This supervision preserves the semantic content of $y$ while remaining within the model's native distribution, allowing for effective continual learning.

\begin{figure*}[t]
  \centering
  \includegraphics[width=\linewidth]{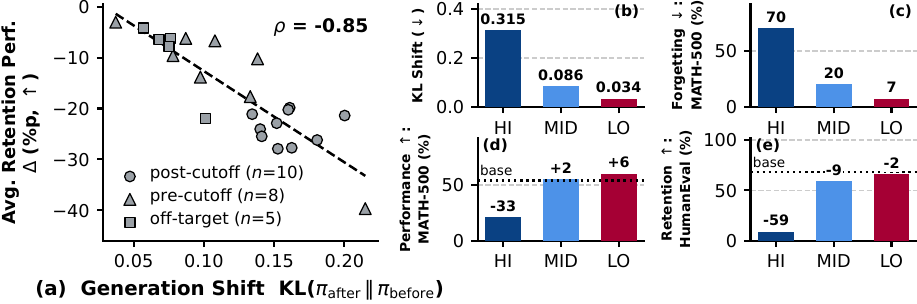}

\caption{\textbf{Distribution shift negatively correlates with ideal continual learning.} \textbf{(a) Experiment 1:} Post-training distribution shift strongly correlates with forgetting existing capabilities across GPQA-Diamond, ARC-C, MMLU-Pro, HumanEval, and MBPP, regardless of data source or type. \textbf{(b) - (e) Experiment 2:} Transforming the same dataset from its original OOD form (HI) to more in-distribution forms (MID, LO) monotonically improves \textbf{(c)} target domain (math) retention, \textbf{(d)} target domain performance, and \textbf{(e)} cross-domain (coding) retention.}
  \vspace{-0.5em}
  \label{fig:shift}
\end{figure*}

\subsection{In-distribution Demonstration Injection}

\subsubsection{In-distribution learning is key for continual learning}\label{sec:dist_cont}

Building on the observation that SDFT preserves the native distribution, we hypothesize that training on out-of-distribution (OOD) targets is the primary driver of poor acquisition and forgetting. To test this, we continually fine-tune \texttt{Qwen2.5-3B-Instruct} \citep{qwen2025qwen25technicalreport}, measuring the resulting policy shift via $D_{\mathrm{KL}}(\pi_{\text{after}} \,\|\, \pi_{\text{before}})$. To avoid data contamination, we use this older model to evaluate against post-release datasets. Further details are available in Appendix \ref{app:id-setup}.

\paragraph{Experiment 1: Distribution shift correlates with forgetting across datasets.} \textbf{[Setup]} First, we collect 23 diverse math training datasets, size-matched to 2.5M completion tokens. Datasets are grouped by domain and release date relative to the base model: \textbf{(1)} pre-cutoff, \textbf{(2)} post-cutoff: long chain-of-thought (CoT) math reasoning, and \textbf{(3)} off-target: post-cutoff code or general instruction data.\footnote{Although a dataset's release date does not definitively guarantee whether its tokens were seen during training, it serves as a reasonable proxy for distinguishing between older and newer data. Our overall results and analyses remain robust regardless of this specific grouping.} Forgetting is measured as the mean performance drop ($\Delta$) across five non-math benchmarks: GPQA-Diamond \citep{rein2024gpqa}, ARC-C \citep{clark2018think}, MMLU-Pro \citep{wang2024mmlupro}, HumanEval \citep{chen2021codex}, and MBPP \citep{50670}. \textbf{[Results]} Fig.~\ref{fig:shift} (a) demonstrates that the magnitude of forgetting (y-axis) is strongly correlated with the distribution shift of the policy $\pi$ after fine-tuning (Spearman $\rho = -0.85$), suggesting that updates causing significant distribution shifts are detrimental to continual learning.

\paragraph{Experiment 2: Distribution shift correlates with poor acquisition and forgetting within a dataset.} \textbf{[Setup]} Second, to isolate distribution shift within an underlying dataset, we hold the corpus fixed using OpenR1-Math-220k \citep{openr1}, a long-CoT dataset. Using identical prompts, we construct three 2M completion-token datasets that vary only in target text shift: (\textbf{HI}) high shift: full R1 trace with reasoning, (\textbf{MID}) mid shift: R1 final solution without long-CoT, and (\textbf{LO}) low shift: the model's own verified-correct solution. LO is curated via self-rejection sampling \citep{singh2024beyond}. The base model decodes a single solution per problem under the same prompt, which is then verified for correctness. \textbf{[Results]} Fig.~\ref{fig:shift} (c) - (e) show that higher distribution shift tracks with in-domain forgetting and catastrophic forgetting, whereas low-shift self-generated samples improve math acquisition performance while preserving existing coding capabilities.

\subsection{Limitations of Self-distillation} 

\begin{figure*}[bt]
\centering
\begin{subfigure}[b]{0.49\textwidth}
  \centering
  \includegraphics[width=\linewidth]{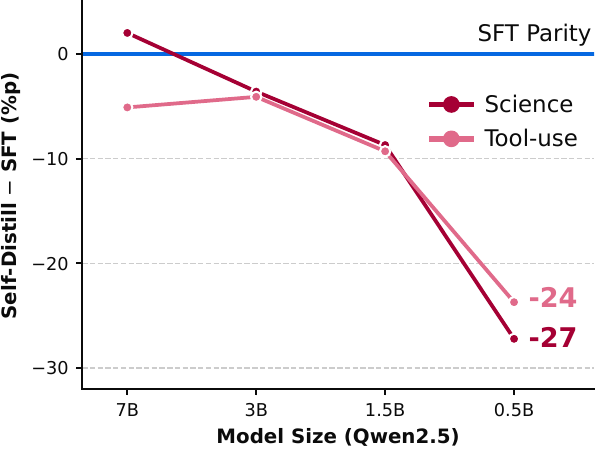}
  \subcaption{SDFT acquisition performance relative to SFT declines monotonically as model size decreases.}
  \label{fig:sdft-scale}
\end{subfigure}
\hfill
\begin{subfigure}[b]{0.49\textwidth}
  \centering
  \includegraphics[width=\linewidth]{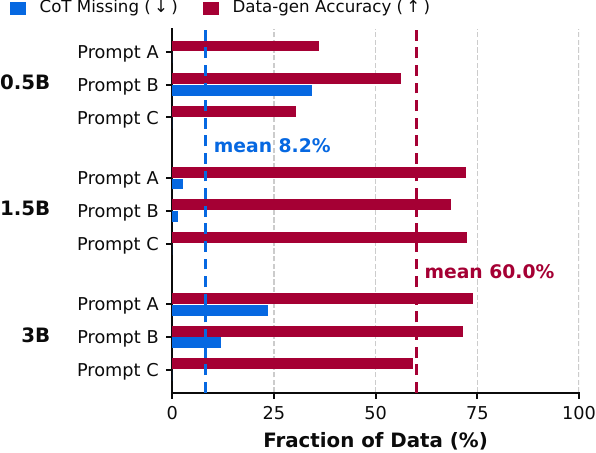}
\subcaption{For example, in the tool-use domain, self-rewriting results in reduced accuracy and missing CoTs.}
  \label{fig:selfrewrite-tooluse}
\end{subfigure}
\caption{\textbf{SDFT fails at smaller model sizes as its ability to self-teach deteriorates.}}
\label{fig:sdft-selfrewrite}
\vspace{-0.5em}
\end{figure*}

\paragraph{Self-supervision fails at small model scales.} 
Although on-policy self-distillation (SDFT; \cite{shenfeld2026selfdistillation}) successfully preserves the base distribution during continual post-training, its efficacy drops for smaller models. Fig.~\ref{fig:sdft-scale} demonstrates that SDFT's performance relative to SFT decreases monotonically with student size. We attribute this to poor self-supervision quality. To evaluate this, we generate an offline dataset to approximate on-policy teacher quality. Fig.~\ref{fig:selfrewrite-tooluse} confirms that smaller models yield weaker supervision: regardless of the prompt (Appendix \ref{app:prompts}), average accuracy falls to 60\%, and CoT traces are dropped 8.2\% of the time.

\subsection{Off-policy In-distribution distillation as Synthetic Self-distillation}\label{sec:synthself}

To bridge this gap, \tsc{SynthSelf} is designed to \textbf{(1)} remain in-distribution with respect to the base model $\pi$, and \textbf{(2)} provide high-quality supervision. We achieve this via two core mechanisms: \textbf{(i)} utilizing off-policy distillation to enable rejection sampling, and \textbf{(ii)} selecting a larger aligner model that simultaneously \textbf{(ii.a.)} yields high-quality supervision, \textbf{(ii.b.)} stays in-distribution relative to the base model, and \textbf{(ii.c.)} incurs low computational overhead.

\paragraph{(i) Off-policy supervision enables rejection sampling.} 
By generating offline datasets for SFT rather than training on-policy, we can filter for accuracy through rejection sampling. This step is essential because small models are highly sensitive to supervision quality.

\paragraph{(ii) Identifying a low-cost, competent in-distribution aligner.}
We formalize the dual objectives of \textbf{(ii.a.)} generating high-quality supervision and \textbf{(ii.b.)} maintaining in-distribution alignment. Let $\mathcal{A}=\{\pi_A^{(1)},\dots,\pi_A^{(K)}\}$ be a pool of candidate aligners. We aim to select an aligner $\pi_A^\star \in \mathcal{A}$ that minimizes the distributional gap to the base model, subject to quality constraints (Eq. \ref{eq:aligner-select}). 

\begin{figure*}[bt!]
    \centering
    \begin{minipage}[t]{0.49\textwidth}
        \centering
        \includegraphics[width=\textwidth]{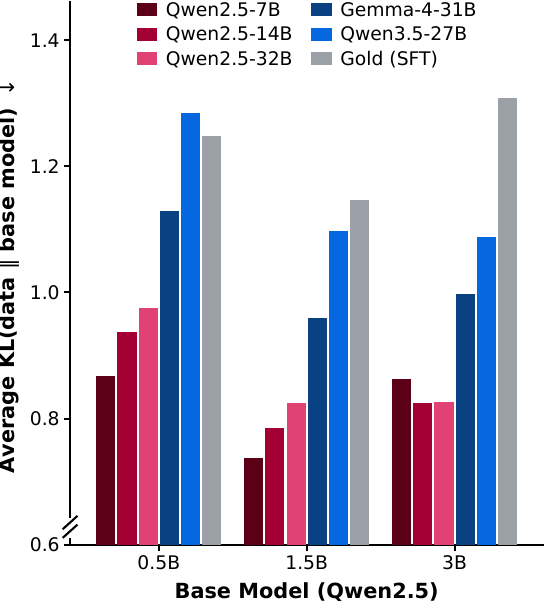}
        \caption{\textbf{Same-family rewrites provide more in-distribution off-policy data relative to Gold SFT.} Conversely, stronger, cross-family models fail to meaningfully improve distributional alignment with the base model.}
        \label{fig:teacher_dist}
    \end{minipage}
    \hfill
    \begin{minipage}[t]{0.49\textwidth}
        \centering
        \includegraphics[width=\textwidth]{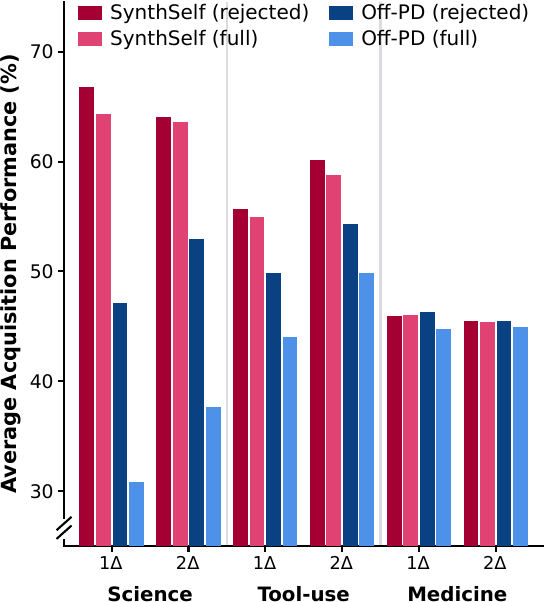}
        \caption{\textbf{An ablation study demonstrates that off-policy rejection sampling is a valuable component.} Furthermore, \tsc{SynthSelf} consistently outperforms standard off-policy distillation, even when controlling for model scale.}
        \label{fig:ablation}
    \end{minipage}
    \vspace{-0.5em}
\end{figure*}

Given a demonstration pair $(x_i,y_i)$ and a rewrite prompt $P^{\text{rewrite}}$ (Box \ref{box:prompt_a}), the aligner's rewrite distribution is:
\begin{equation}
q_i(y)\;\coloneqq\;\pi_A\!\left(y \mid P^{\text{rewrite}}(x_i, y_i)\right).
\label{eq:rewrite}
\end{equation}
This distribution translates the knowledge from $(x_i,y_i)$ into the aligner's native distribution. Appendix \ref{app:samples} shows an example: the rewrite keeps the gold answer and format while restating the reasoning in the aligner's own phrasing.
We decode greedily, $\hat{y}_{i,t}=\arg\max_{v\in\mathcal{V}}\,\pi_A\!\left(v \mid P^{\text{rewrite}}(x_i,y_i),\,\hat{y}_{i,<t}\right)$, ensuring a deterministic rewrite $\hat{y}_i$. We measure supervision quality \textbf{(ii.a.)} directly on these rewrites:
\begin{equation}
R(\pi_A)\;=\;\mathbb{E}_{i}\big[r(x_i,\hat{y}_i)\big],
\qquad
\hat{y}_i\;=\;\operatorname{greedy}(q_i),
\label{eq:reward}
\end{equation}
where a verifiable reward $r(x_i, \hat{y}_i) = \mathbb{I}[a(\hat{y}_i) = a(y_i)]$ checks the extracted answer $a(\cdot)$ against the gold demonstration. 

The student is trained to imitate the point mass $\delta_{\hat{y}_i}$ at the mode of $q_i$. The divergence against this point mass simplifies to the negative log-likelihood (NLL) under the student $\pi$, serving as our metric for in-distribution alignment \textbf{(ii.b.)}:
\begin{equation}
\mathcal{L}(\pi_A)\;=\;\mathbb{E}_{i}\big[D_{\mathrm{KL}}\big(\delta_{\hat{y}_i} \,\|\, \pi(\cdot\mid x_i)\big)\big]
\;=\;-\,\mathbb{E}_{i}\sum_{t=1}^{|\hat{y}_i|}\log\pi\big(\hat{y}_{i,t}\mid x_i,\hat{y}_{i,<t}\big).
\label{eq:aligner-div}
\end{equation}

Standard SFT on gold demonstrations follows an identical objective, replacing $\delta_{\hat{y}_i}$ with $\delta_{y_i}$ to provide a baseline divergence:
\begin{equation}
\mathcal{L}_{\text{gold}}\;=\;\mathbb{E}_{i}\big[D_{\mathrm{KL}}\big(\delta_{y_i}\,\|\,\pi(\cdot\mid x_i)\big)\big]
\;=\;-\,\mathbb{E}_{i}\sum_{t=1}^{|y_i|}\log\pi\big(y_{i,t}\mid x_i,y_{i,<t}\big).
\label{eq:gold-div}
\end{equation}

Consequently, we formulate aligner selection as:
\begin{equation}
\pi_A^{\star}\;=\;\operatorname*{arg\,min}_{\pi_A\in\mathcal{A}}\;\mathcal{L}(\pi_A)
\qquad\text{s.t.}\qquad \mathcal{L}(\pi_A)\;<\;\mathcal{L}_{\text{gold}}
\quad\text{and}\quad
R(\pi_A)\;>\;R(\pi).
\label{eq:aligner-select}
\end{equation}
In short, we select the aligner closest in distribution to the student. The constraints guarantee its rewrites offer a distributional advantage over standard SFT ($\mathcal{L}(\pi_A) < \mathcal{L}_{\text{gold}}$) and strictly exceed the base model's supervision quality ($R(\pi_A) > R(\pi)$).

In practice, scaling up within the same model family perfectly satisfies Eq.~\ref{eq:aligner-select}. Fig.~\ref{fig:teacher_dist} shows that same-family aligners (\ttt{Qwen2.5-7B}, \ttt{14B}, \ttt{32B}) achieve the lowest average KL, falling well below $\mathcal{L}_{\text{gold}}$, whereas stronger cross-family models fail to meaningfully reduce the NLL (averaged across three domains detailed in \S\ \ref{sec:exp}). Furthermore, same-family aligners yield higher data accuracy than the base model, satisfying $R(\pi_A) > R(\pi)$ (Tab.~\ref{tab:aligner-family}). Finally, to satisfy the low-cost constraint \textbf{(ii.c.)}, we instantiate $\pi_A^\star$ using minimally larger \ttt{Qwen2.5-7B} and \ttt{14B} models, ensuring minimal computational overhead.

\begin{algorithm2e}[t]
    \caption{\tsc{SynthSelf} (post-aligner selection)}
    \label{alg:synthself}
    \SetAlgoLined
    \LinesNumbered
    \Input{Demonstrations $\mc{D}_\text{train} = \{(x_i, y_i)\}_{i=1}^N$, post-trained base model $\pi_{\theta}$, aligner $\pi_A^{\star}$, rewrite instruction $P^\text{rewrite}$, answer extractor $a(\cdot)$}

    $\hat{\mc{D}} \leftarrow \emptyset$ \tcp*{Initialize offline dataset}

    \ForEach{$(x_i, y_i) \in \mc{D}_\text{train}$}{
        \tcc{\textcolor{LGRedDark}{\textbf{Synthetic In-distribution Rewrite}}}
        $\hat{y}_i \leftarrow \arg\max_{y}\; \pi_A\!\left(y \mid P^\text{rewrite}(x_i, y_i)\right)$ \tcp*{Greedy decoding}

        \tcc{\textcolor{LGRed}{\textbf{Rejection Sampling}}}
        \eIf{$a(\hat{y}_i) = a(y_i)$}{
    $\hat{\mc{D}} \leftarrow \hat{\mc{D}} \cup \{(x_i, \hat{y}_i)\}$ \tcp*{Same answer, but in-distribution}
        }{
            \textbf{discard} $(x_i, \hat{y}_i)$
        }
    }
    \tcc{\textcolor{LGRedLight}{\textbf{Off-policy learning via SFT}}}
    $\hat{\theta} \leftarrow \arg\min_{\theta}\; \mathbb{E}_{(x, \hat{y}) \sim \hat{\mc{D}}}\!\left[-\log \pi_{\theta}(\hat{y} \mid x)\right]$ \;

    \Return $\pi_{\hat{\theta}}$ \;
\end{algorithm2e}

\paragraph{Algorithmic implementation.} Using this minimally larger same-family aligner $\pi_A$, we implement \tsc{SynthSelf} (see Alg. \ref{alg:synthself}). We first initialize an empty offline dataset $\hat{\mc{D}}$ (Line 1). Next, processing the data in batches, we generate an in-distribution rewrite $\hat{y}_i$ for each sample $i$ via the aligner $\pi_A$ using greedy decoding (Lines 2 and 3). Samples that satisfy the rejection sampling are then added to $\hat{\mc{D}}$ (Lines 4 to 8). Finally, we perform SFT on this filtered dataset $\hat{\mc{D}}$ to update the base model parameters (Line 10).

\section{Empirical Study}\label{sec:exp}

We evaluate \tsc{SynthSelf} on the three domains from SDFT \citep{shenfeld2026selfdistillation}, alongside a 4-subject science mixture to verify generalization. In \S\ref{sec:further}, we further analyze our approach by: \textbf{(i)} demonstrating that distribution matching outweighs aligner capability; \textbf{(ii)} benchmarking against SDFT on matched and diverse prompts; and \textbf{(iii)} ablating rejection sampling to confirm the value of off-policy supervision.

\subsection{Experimental Protocol} \label{sec:protocol}

\definecolor{rowbase}{HTML}{F0F1F3}
\definecolor{rowours}{HTML}{333333}
\definecolor{rowtask}{HTML}{D8DBE0}
\begin{table*}[bt!]
\centering
\setlength{\tabcolsep}{3.5pt}
\resizebox{\textwidth}{!}{
\begin{tabular}{llrrrrrrrrrrr}
\toprule
 & & \multicolumn{5}{c}{\textbf{Baseline}} & \multicolumn{6}{c}{\textbf{\tsc{SynthSelf} (ours)}} \\
\cmidrule(lr){3-7} \cmidrule(lr){8-13}
\textbf{Model} & & Base & SFT & SDFT & DFT & ASFT & \tsc{SS}-1$\Delta$ & $\Delta$ Off-PD & $\Delta$ On-PD & \tsc{SS}-2$\Delta$ & $\Delta$ Off-PD & $\Delta$ On-PD \\
\midrule
\rowcolor{rowtask}\multicolumn{13}{c}{\tsc{\textbf{Science: Chemistry}}} \\
\midrule
0.5B & \textbf{Acq.} & 25.4 & 56.8 & 38.5 & 35.5 & 45.6 & 63.7 & +21.1 & +34.5 & 59.4 & +10.1 & +21.5 \\
 & \textbf{Ret.} & 31.8 & 28.2 & 30.3 & 28.7 & 30.3 & 27.8 & -2.4 & +1.2 & 29.5 & -0.8 & +5.1 \\
\addlinespace[2pt]
1.5B & \textbf{Acq.} & 31.0 & 64.9 & 53.6 & 56.8 & 48.5 & 67.1 & +16.4 & +33.3 & 65.3 & +12.0 & +22.9 \\
 & \textbf{Ret.} & 43.9 & 44.1 & 41.8 & 43.8 & 45.2 & 45.0 & +0.4 & +1.9 & 44.6 & +0.0 & +2.5 \\
\addlinespace[2pt]
3B & \textbf{Acq.} & 30.2 & 68.8 & 65.9 & 60.9 & 58.6 & 69.6 & +21.5 & +38.7 & 67.7 & +11.2 & +21.9 \\
 & \textbf{Ret.} & 52.9 & 52.6 & 51.5 & 51.3 & 51.6 & 52.7 & -0.7 & +2.9 & 51.8 & -2.3 & -1.3 \\
\addlinespace[2pt]
\textbf{Avg.} & \textbf{Acq.} & 28.9 & \cellcolor[HTML]{C0DFC4}63.5 & \cellcolor[HTML]{F2B9B4}52.7 & \cellcolor[HTML]{F2ABA6}51.1 & \cellcolor[HTML]{F2A9A4}50.9 & \cellcolor[HTML]{9ED4A8}66.8 & \textcolor[HTML]{1B7A45}{+19.7} & \textcolor[HTML]{1B7A45}{+35.5} & \cellcolor[HTML]{BADDBF}64.1 & \textcolor[HTML]{1B7A45}{+11.1} & \textcolor[HTML]{1B7A45}{+22.1} \\
 & \textbf{Ret.} & 42.8 & \cellcolor[HTML]{F1DBD8}41.6 & \cellcolor[HTML]{F2A9A4}41.2 & \cellcolor[HTML]{F2ADA8}41.2 & \cellcolor[HTML]{9ED4A8}42.4 & \cellcolor[HTML]{EAEDE7}41.8 & \textcolor[HTML]{B03A2E}{-0.9} & \textcolor[HTML]{1B7A45}{+2.0} & \cellcolor[HTML]{D5E6D6}42.0 & \textcolor[HTML]{B03A2E}{-1.0} & \textcolor[HTML]{1B7A45}{+2.1} \\
\midrule
\rowcolor{rowtask}\multicolumn{13}{c}{\tsc{\textbf{Science Mixture (4 subjects)}}} \\
\midrule
0.5B & \textbf{Acq.} & 23.0 & 48.7 & 25.7 & 32.8 & 35.5 & 49.8 & +7.2 & +20.0 & 43.0 & +0.4 & +9.6 \\
 & \textbf{Ret.} & 31.8 & 30.0 & 29.5 & 29.1 & 29.7 & 30.8 & -0.7 & +2.4 & 28.8 & -2.5 & +1.8 \\
\addlinespace[2pt]
1.5B & \textbf{Acq.} & 30.4 & 55.5 & 40.2 & 52.8 & 48.9 & 58.9 & +12.6 & +18.7 & 57.0 & +7.0 & +14.3 \\
 & \textbf{Ret.} & 43.9 & 42.8 & 43.1 & 43.3 & 44.3 & 43.3 & -1.5 & -1.1 & 44.5 & -2.3 & +0.9 \\
\addlinespace[2pt]
3B & \textbf{Acq.} & 43.8 & 61.1 & 60.6 & 50.9 & 55.1 & 63.2 & +7.2 & +17.4 & 67.7 & +10.4 & +16.6 \\
 & \textbf{Ret.} & 52.9 & 52.9 & 51.5 & 50.7 & 51.6 & 51.6 & -2.5 & +1.4 & 52.9 & +0.0 & +2.0 \\
\addlinespace[2pt]
\textbf{Avg.} & \textbf{Acq.} & 32.4 & \cellcolor[HTML]{B6DCBC}55.1 & \cellcolor[HTML]{F2A9A4}42.2 & \cellcolor[HTML]{F1C7C3}45.5 & \cellcolor[HTML]{F1D1CD}46.5 & \cellcolor[HTML]{9ED4A8}57.3 & \textcolor[HTML]{1B7A45}{+9.0} & \textcolor[HTML]{1B7A45}{+18.7} & \cellcolor[HTML]{ADD9B5}55.9 & \textcolor[HTML]{1B7A45}{+6.0} & \textcolor[HTML]{1B7A45}{+13.5} \\
 & \textbf{Ret.} & 42.8 & \cellcolor[HTML]{BBDEC0}41.9 & \cellcolor[HTML]{F1D4D0}41.3 & \cellcolor[HTML]{F2A9A4}41.0 & \cellcolor[HTML]{BFDFC3}41.9 & \cellcolor[HTML]{BCDEC1}41.9 & \textcolor[HTML]{B03A2E}{-1.6} & \textcolor[HTML]{1B7A45}{+0.9} & \cellcolor[HTML]{9ED4A8}42.1 & \textcolor[HTML]{B03A2E}{-1.6} & \textcolor[HTML]{1B7A45}{+1.5} \\
\midrule
\rowcolor{rowtask}\multicolumn{13}{c}{\tsc{\textbf{Tool-use}}} \\
\midrule
0.5B & \textbf{Acq.} & 23.7 & 40.2 & 24.7 & 30.9 & 42.3 & 43.3 & -1.0 & +7.2 & 56.7 & +10.3 & +6.2 \\
 & \textbf{Ret.} & 31.8 & 25.9 & 30.5 & 26.0 & 29.0 & 28.3 & -1.0 & +0.5 & 29.9 & -0.5 & +3.2 \\
\addlinespace[2pt]
1.5B & \textbf{Acq.} & 43.3 & 57.7 & 51.5 & 59.8 & 54.6 & 61.9 & +11.3 & +15.5 & 55.7 & -2.1 & -6.2 \\
 & \textbf{Ret.} & 43.9 & 39.9 & 42.6 & 41.6 & 41.6 & 42.3 & -1.1 & -3.0 & 42.4 & -1.4 & +0.3 \\
\addlinespace[2pt]
3B & \textbf{Acq.} & 33.0 & 66.0 & 57.7 & 47.4 & 58.8 & 61.9 & +7.2 & +13.4 & 68.0 & +9.3 & +9.3 \\
 & \textbf{Ret.} & 52.9 & 50.9 & 52.2 & 49.6 & 53.2 & 51.0 & +0.5 & +0.9 & 51.4 & +0.4 & +2.1 \\
\addlinespace[2pt]
\textbf{Avg.} & \textbf{Acq.} & 33.3 & \cellcolor[HTML]{D8E7D8}54.6 & \cellcolor[HTML]{F2A9A4}44.7 & \cellcolor[HTML]{F2B5B1}46.0 & \cellcolor[HTML]{F0EAE7}51.9 & \cellcolor[HTML]{CDE4CF}55.7 & \textcolor[HTML]{1B7A45}{+5.8} & \textcolor[HTML]{1B7A45}{+12.0} & \cellcolor[HTML]{9ED4A8}60.1 & \textcolor[HTML]{1B7A45}{+5.8} & \textcolor[HTML]{1B7A45}{+3.1} \\
 & \textbf{Ret.} & 42.8 & \cellcolor[HTML]{F2A9A4}38.9 & \cellcolor[HTML]{9ED4A8}41.8 & \cellcolor[HTML]{F2B1AC}39.1 & \cellcolor[HTML]{BBDEC0}41.3 & \cellcolor[HTML]{E5EBE3}40.5 & \textcolor[HTML]{B03A2E}{-0.5} & \textcolor[HTML]{B03A2E}{-0.5} & \cellcolor[HTML]{BDDEC1}41.2 & \textcolor[HTML]{B03A2E}{-0.5} & \textcolor[HTML]{1B7A45}{+1.8} \\
\midrule
\rowcolor{rowtask}\multicolumn{13}{c}{\tsc{\textbf{Medicine}}} \\
\midrule
0.5B & \textbf{Acq.} & 26.7 & 16.0 & 26.7 & 2.0 & 18.2 & 32.7 & -3.0 & -1.5 & 34.2 & -0.5 & -2.0 \\
 & \textbf{Ret.} & 31.8 & 28.4 & 30.5 & 28.0 & 29.4 & 31.0 & -0.8 & +3.7 & 29.9 & -2.4 & +2.1 \\
\addlinespace[2pt]
1.5B & \textbf{Acq.} & 42.8 & 41.3 & 28.8 & 27.7 & 36.7 & 47.8 & +0.5 & -0.2 & 48.3 & -0.3 & -3.0 \\
 & \textbf{Ret.} & 43.9 & 41.4 & 42.5 & 40.8 & 42.8 & 43.7 & -2.8 & +4.5 & 43.3 & -2.6 & +2.2 \\
\addlinespace[2pt]
3B & \textbf{Acq.} & 57.5 & 48.8 & 54.8 & 20.7 & 50.2 & 57.3 & +1.3 & +5.2 & 54.0 & +1.0 & +0.7 \\
 & \textbf{Ret.} & 52.9 & 48.7 & 52.2 & 51.7 & 52.8 & 52.3 & -3.0 & +1.4 & 52.8 & -0.6 & -0.6 \\
\addlinespace[2pt]
\textbf{Avg.} & \textbf{Acq.} & 42.3 & \cellcolor[HTML]{D9E8D9}35.4 & \cellcolor[HTML]{D2E5D3}36.8 & \cellcolor[HTML]{F2A9A4}16.8 & \cellcolor[HTML]{DCE8DB}35.0 & \cellcolor[HTML]{9ED4A8}45.9 & \textcolor[HTML]{B03A2E}{-0.4} & \textcolor[HTML]{1B7A45}{+1.2} & \cellcolor[HTML]{A0D5AA}45.5 & \textcolor[HTML]{1B7A45}{+0.1} & \textcolor[HTML]{B03A2E}{-1.4} \\
 & \textbf{Ret.} & 42.8 & \cellcolor[HTML]{F2A9A4}39.5 & \cellcolor[HTML]{C3E0C6}41.7 & \cellcolor[HTML]{F1C9C5}40.2 & \cellcolor[HTML]{C7E1CA}41.7 & \cellcolor[HTML]{9ED4A8}42.4 & \textcolor[HTML]{B03A2E}{-2.2} & \textcolor[HTML]{1B7A45}{+3.2} & \cellcolor[HTML]{B1DAB8}42.0 & \textcolor[HTML]{B03A2E}{-1.9} & \textcolor[HTML]{1B7A45}{+1.2} \\
\midrule
\rowcolor{rowours}\multicolumn{13}{c}{\textcolor{white}{\tsc{\textbf{Average}}}} \\
\midrule
\textbf{Avg.} & \textbf{Acq.} & 34.2 & \cellcolor[HTML]{C8E2CB}52.2 & \cellcolor[HTML]{F1CDC9}44.1 & \cellcolor[HTML]{F2A9A4}39.8 & \cellcolor[HTML]{F0DEDA}46.1 & \cellcolor[HTML]{9ED4A8}56.4 & \textcolor[HTML]{1B7A45}{+8.5} & \textcolor[HTML]{1B7A45}{+16.9} & \cellcolor[HTML]{9ED4A8}56.4 & \textcolor[HTML]{1B7A45}{+5.7} & \textcolor[HTML]{1B7A45}{+9.3} \\
 & \textbf{Ret.} & 42.8 & \cellcolor[HTML]{F2B4B0}40.5 & \cellcolor[HTML]{C2E0C6}41.5 & \cellcolor[HTML]{F2A9A4}40.4 & \cellcolor[HTML]{A3D6AC}41.8 & \cellcolor[HTML]{B2DBB9}41.7 & \textcolor[HTML]{B03A2E}{-1.3} & \textcolor[HTML]{1B7A45}{+1.4} & \cellcolor[HTML]{9ED4A8}41.8 & \textcolor[HTML]{B03A2E}{-1.2} & \textcolor[HTML]{1B7A45}{+1.7} \\
\bottomrule
\end{tabular}
}
\caption{\textbf{\tsc{SynthSelf} leads acquisition (Acq.) in all 4 benchmarks and ties the best baseline on 14 retention tasks (Ret.), over 4 baselines, off-/on-policy distillation, and three base models.} \colorbox[HTML]{9ED4A8}{Green} and \colorbox[HTML]{F2A9A4}{red} background shading on Avg. rows is min-max normalized across each row. The 14 retention tasks span instruction following, commonsense, diverse knowledge, code, and math. The mixture contains Chemistry, Biology, Materials, and Physics in both the train and test splits. $\Delta$ Off-PD and $\Delta$ On-PD are \tsc{SynthSelf} minus off- or on-policy distillation (\%p).}
\label{tab:main}
    \vspace{-0.5em}
\end{table*}

\paragraph{Baselines.} We compare our method against four baselines: standard SFT, SDFT \citep{shenfeld2026selfdistillation}, DFT \citep{wu2026on}, which modifies the objective function to imitate on-policy learning, naturally inducing in-distribution learning, and ASFT \citep{zhu2026anchored}, which improves upon DFT by further mitigating distributional shift. To control for the larger aligner model size, we also compare \tsc{SynthSelf} against off- and on-policy distillation \citep{agarwal2024onpolicy}. Here, off-policy distillation is equivalent to Alg. \ref{alg:synthself}, but conditions the generation on the original query $x_i$ rather than the rewrite prompt $P^\text{rewrite}(x_i, y_i)$. We tune standard SFT minimally and fix these hyperparameters across all methods for fairness. To ensure a strong baseline, we additionally optimize SDFT's skip-token hyperparameter; which discards $T$ initial formatting tokens. Conversely, \tsc{SynthSelf} is tuning-free and introduces zero additional hyperparameters.

\paragraph{Datasets and models.} Following \citet{shenfeld2026selfdistillation}, we evaluate on three domains: science (SciKnowEval Chemistry L-3; \citet{feng2025sciknowevalevaluatingmultilevelscientific}), tool-use (ToolAlpaca; \citet{tang2023toolalpacageneralizedtoollearning}), and medicine (MedMCQA; \citet{pmlr-v174-pal22a}). We expand to all four SciKnowEval subjects for our mixture experiment. We substitute the medical dataset used by \citet{shenfeld2026selfdistillation} with MedMCQA to avoid LLM-as-a-judge evaluation and improve reproducibility. We adopt the \ttt{Qwen2.5-Instruct} family as base models, unlike later versions like \ttt{Qwen3} \citep{yang2025qwen3technicalreport} exhaustively trained on open-source data; this choice reduces contamination and ensures demonstration novelty. Further experimental details are in Appendix \ref{app:exp}.

\subsection{Results}\label{sec:results}

\tsc{SynthSelf} achieves the highest acquisition across all four benchmarks, including the science mixture, while matching the strongest baselines on 14 retention tasks (Tab.~\ref{tab:main}, Fig.~\ref{fig:headline}). We abbreviate \tsc{SynthSelf} as \tsc{SS}, with 1$\Delta$ and 2$\Delta$ denoting aligners one and two sizes larger, respectively (i.e., \ttt{Qwen2.5-7B} and \ttt{Qwen2.5-14B}). Interestingly, recent methods fail to surpass SFT in acquisition at these smaller scales. DFT's performance in the medical domain deteriorates especially significantly; we have double-checked that these evaluation values are correct. However, consistent with the discussion in \S\ \ref{sec:dist_cont}, the on-policy and in-distribution mechanisms of SDFT and ASFT successfully improve retention over SFT. 

\definecolor{rowours}{HTML}{333333}
\definecolor{rowtask}{HTML}{D8DBE0}
\begin{table*}[bt!]
\centering
\setlength{\tabcolsep}{4pt}
\small
\resizebox{\textwidth}{!}{
\begin{tabular}{lrrrrrrrr}
\toprule
 & \multicolumn{2}{c}{\textbf{Acquisition} $\uparrow$} & \multicolumn{2}{c}{\textbf{KL (NLL)} $\downarrow$} & \multicolumn{2}{c}{\textbf{Data-gen Accuracy (\%)} $\uparrow$} & \multicolumn{2}{c}{\textbf{Data-gen PFLOPs} $\downarrow$} \\
\cmidrule(lr){2-3} \cmidrule(lr){4-5} \cmidrule(lr){6-7} \cmidrule(lr){8-9}
\textbf{Aligner} & \tsc{Sci.} & \tsc{Tool} & \tsc{Sci.} & \tsc{Tool} & \tsc{Sci.} & \tsc{Tool} & \tsc{Sci.} & \tsc{Tool} \\
\midrule
\rowcolor{rowtask}\multicolumn{9}{c}{\textbf{More In-distribution Aligners}} \\
\midrule
\quad \ttt{Qwen2.5-7B} & \textbf{66.8} & 55.7 & \textbf{0.995} & \textbf{0.650} & 99.0 & 84.1 & \textbf{13.9} & \textbf{39.0} \\
\quad \ttt{Qwen2.5-14B} & 64.1 & \textbf{60.1} & 1.033 & 0.666 & 99.1 & 81.8 & 36.6 & 76.0 \\
\midrule
\rowcolor{rowtask}\multicolumn{9}{c}{\textbf{Stronger but More Out-of-distribution Aligners}} \\
\midrule
\quad \ttt{Qwen3.5-27B} & 65.1 & 50.5 & 1.440 & 0.873 & 90.3 & 95.8 & 70.0 & 142.1 \\
\quad \ttt{Gemma-4-31B} & 60.8 & 56.0 & 1.245 & 0.812 & \textbf{99.8} & \textbf{99.3} & 148.2 & 167.9 \\
\midrule
\rowcolor{rowours}\multicolumn{9}{c}{\textcolor{white}{\textbf{Average}}} \\
\midrule
\quad ID & 65.5 & 57.9 & 1.014 & 0.658 & 99.0 & 82.9 & 25.2 & 57.5 \\
\quad OOD & 63.0 & 53.3 & 1.342 & 0.843 & 95.0 & 97.5 & 109.1 & 155.0 \\
\quad $\Delta$ = ID $-$ OOD & \textcolor[HTML]{1B7A45}{+2.5} & \textcolor[HTML]{1B7A45}{+4.6} & \textcolor[HTML]{1B7A45}{-0.328} & \textcolor[HTML]{1B7A45}{-0.185} & \textcolor[HTML]{1B7A45}{+4.0} & \textcolor[HTML]{B03A2E}{-14.6} & \textcolor[HTML]{1B7A45}{-83.8} & \textcolor[HTML]{1B7A45}{-97.5} \\
\bottomrule
\end{tabular}
}
\caption{\textbf{\tsc{SynthSelf} performs better with in-distribution aligners.} Averaged over the
\ttt{Qwen2.5-0.5B}, \ttt{1.5B}, and \ttt{3B} base models, using aligners that are more in-distribution (measured by NLL) is a stronger predictor of performance than the strength of the aligner (measured by data-generation accuracy). Data-generation accuracy represents the reward $R(\pi_A)$ in Eq. \ref{eq:reward}. It is additionally advantageous that smaller, same-family aligners are the most compute-efficient when
generating off-policy data.
\textbf{Bold} indicates best value across each column.}
\label{tab:aligner-family}
    \vspace{-0.5em}
\end{table*}

Moreover, we validate that the performance gains of \tsc{SynthSelf} stem primarily from
in-distribution demonstration injection, rather than merely leveraging a larger aligner
(Tab.~\ref{tab:main}). Off- and on-policy distillation (Off-/On-PD) use the identical
aligner,
differing only in conditioning on $x_i$ rather than on $P^{\text{rewrite}}(x_i, y_i)$. Averaged across all four settings, \tsc{SynthSelf} nonetheless improves acquisition by
+5.7\%p to +16.9\%p while maintaining comparable retention. As expected, demonstration injection drives these acquisition gains, whereas all three methods (\tsc{SynthSelf}, Off-PD, and On-PD) similarly preserve the base model distribution to sustain retention capabilities. Lastly, we observe that \tsc{SynthSelf} does not yield significant improvements over Off-/On-PD in the medical domain. This is likely due to the aligner models already having perfect knowledge of the training data, making demonstration injection largely redundant. Despite this, \tsc{SynthSelf} remains highly competitive, successfully outperforming all remaining baselines.

\subsection{Further Experimental Analyses}\label{sec:further}

\paragraph{Distribution matching outweighs aligner capability, and leads to data generation cost savings.} We validate our hypothesis that lower-cost, in-distribution aligners are more effective than stronger out-of-distribution (OOD) alternatives. Although larger aligners like \ttt{Qwen3.5-27B} and \ttt{Gemma-4-31B} \citep{gemmateam2026gemma4technicalreport} generate higher-quality supervision (measured via data-generation accuracy), they introduce severe distributional shift relative to the base model. Consequently, they yield similar or worse acquisition performance (Tab.~\ref{tab:aligner-family}). Moreover, using a same-family aligner reduces data generation inference costs by 62.9\% to 76.9\% FLOPs.

\begin{figure}
    \centering
    \includegraphics[width=\linewidth]{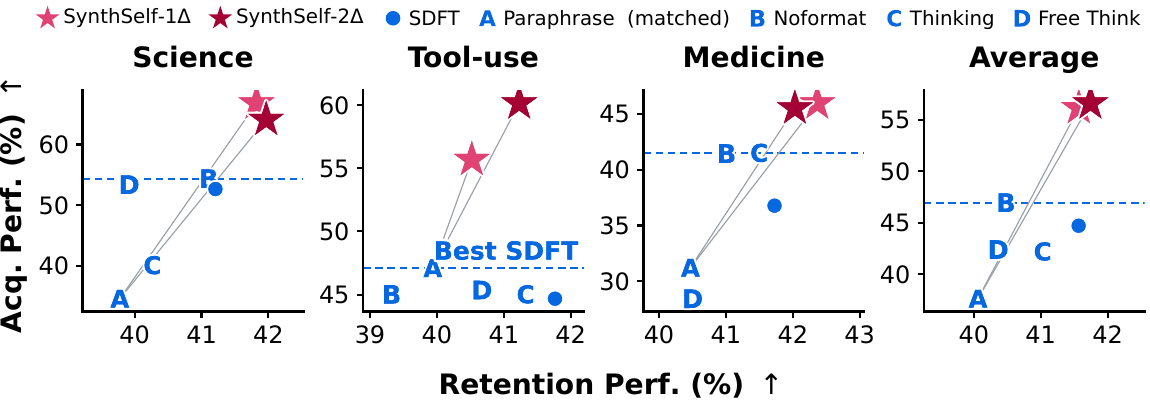}
\caption{\textbf{\tsc{SynthSelf} outperforms SDFT across diverse prompt variations, including a version prompt-matched to \tsc{SynthSelf}.} On average, \tsc{SynthSelf} outperforms prompt-matched SDFT by +18.96\%p and +1.68\%p on acquisition and retention, respectively.}
    \label{fig:sdft_prompt}
    \vspace{-0.5em}
\end{figure}

\paragraph{\tsc{SynthSelf} outperforms prompt-matched self-distillation.} We confirm that \tsc{SynthSelf}'s gains over SDFT do not arise from differences in supervision prompts (Fig.~\ref{fig:sdft_prompt}). Evaluating four SDFT prompt variants, including one matched to \tsc{SynthSelf}, we find our method consistently achieves higher acquisition across all three domains and higher average retention than every variant. Full prompts are provided in Appendix~\ref{app:prompts}.

\paragraph{Rejection sampling ablation.} Finally, we ablate the offline data generation that enables rejection sampling (Alg.~\ref{alg:synthself}, Lines 4--7). As shown in Fig.~\ref{fig:ablation}, this filtering mechanism consistently improves both \tsc{SynthSelf} and Off-PD. We hypothesize such filtering is especially critical for smaller base models, which are highly sensitive to noisy supervision.

\section{Discussion}

\paragraph{Additional related work.} \citet{huang2026how} highlights the impact of distribution drift in SFT using code-generation data on \ttt{Qwen3-8B}. However, they do not directly address continual learning or small language models, restricting their evaluation to limited scales and domains. Furthermore, \citet{li2026raftdatarefinementadaptive} combines mechanisms that we instead isolate: SDFT-style self-rewrites with similarity filtering, cross-family fusion, and an on-policy KL regularizer; covered individually by our SDFT, on-policy distillation, and ASFT baselines. Crucially, we differ in the supervision source. Because their rewrites originate from the student, they inherit the small-scale degradation shown in Fig.~\ref{fig:sdft-selfrewrite}. Our same-family aligner explicitly resolves this limitation.

\paragraph{Training compute savings and further performance headroom.} Rejection sampling reduces training compute by shrinking the dataset $\vert \hat{D} \vert$ (e.g., by 17.1\% for tool use; Tab.~\ref{tab:aligner-family}). Despite training on fewer demonstrations than SFT, \tsc{SynthSelf} improves acquisition by +4.2\%p on average, proving these gains stem from supervision quality rather than quantity. To further boost performance, practitioners can scale $k$ in pass@$k$, trading inference compute for larger coverage supervision.

\paragraph{Limitation and future directions.} \tsc{SynthSelf} requires a larger, same-family aligner. Fortunately, most major open-weight families provide a spectrum of model sizes, naturally satisfying this requirement since smaller variants are often distilled from their larger counterparts (e.g., \ttt{Qwen2.5}, \ttt{Qwen3.5}, \ttt{OLMo 2} \citep{walsh2025}, \ttt{Llama 3} \citep{grattafiori2024llama3herdmodels}, and \ttt{Gemma 4}). Nevertheless, future work could explore adapting out-of-family aligners to produce in-distribution supervision. One promising direction is adding the base model's answer in-context with an appropriate prompt to generate an in-distribution rewrite. 


\subsection*{AI use statement}

We used generative AI to assist with experimental code generation and reviewing writing for
grammar and spelling. Furthermore, language models generated the reasoning demonstrations for three SciKnowEval subjects. The Chemistry demonstrations were generated with \ttt{GPT-4o} by \cite{shenfeld2026selfdistillation}. AI was not used for conceptualization, methodology design, data analysis, or
result interpretation. All reported numbers were produced directly by our experimental pipeline, not
by a language model. The authors reviewed all AI-assisted scripts and text, and we take full
responsibility for the final content of this work.

\newpage
\bibliography{iclr2027_conference}
\bibliographystyle{iclr2027_conference}

\newpage
\appendix
\section*{\textbf{Appendix}}

\section{Prompts} \label{app:prompts}

Prompt A is $P^\text{rewrite}$, the instruction behind every trained \tsc{SynthSelf} cell. Prompts
B, C and D are the rewrite variants used in the self-rewrite study (\S\ \ref{sec:dist_cont}) and the
prompt-matched SDFT ablation (\S\ \ref{sec:further}). \ttt{\{q\}} and \ttt{\{gold\}} are the
question and the gold response.

\begin{tcolorbox}[
    colback=gray!5!white,
    colframe=gray!75!black,
    breakable,
    title={Prompt A: Paraphrase (matched) $=P^\text{rewrite}$},
    label={box:prompt_a}
]
{\ttfamily\small
Rewrite the following response in your own words. Keep the SAME final answer and the SAME format exactly.
\par\smallskip
Question:\\
\{q\}
\par\smallskip
Response to rewrite:\\
\{gold\}
}
\end{tcolorbox}

\begin{tcolorbox}[
    colback=gray!5!white,
    colframe=gray!75!black,
    breakable,
    title={Prompt B: No Format},
    label={box:prompt_b}
]
{\ttfamily\small
Rewrite the following response in your own words. Keep the SAME final answer.
\par\smallskip
Question:\\
\{q\}
\par\smallskip
Response to rewrite:\\
\{gold\}
\par\smallskip
End with exactly this format: \textless{}reasoning\textgreater{}...\textless{}/reasoning\textgreater{}\\
\textless{}answer\textgreater{}LETTER\textless{}/answer\textgreater{}
}
\end{tcolorbox}

\begin{tcolorbox}[
    colback=gray!5!white,
    colframe=gray!75!black,
    breakable,
    title={Prompt C: Thinking},
    label={box:prompt_c}
]
{\ttfamily\small
Rewrite the following response in your own words, including the thinking process. Keep the SAME final answer.
\par\smallskip
Question:\\
\{q\}
\par\smallskip
Response to rewrite:\\
\{gold\}
\par\smallskip
End with exactly this format: \textless{}reasoning\textgreater{}...\textless{}/reasoning\textgreater{}\\
\textless{}answer\textgreater{}LETTER\textless{}/answer\textgreater{}
}
\end{tcolorbox}

\begin{tcolorbox}[
    colback=gray!5!white,
    colframe=gray!75!black,
    breakable,
    title={Prompt D: Free Think},
    label={box:prompt_d}
]
{\ttfamily\small
Rewrite the following response in your own words, adding your thinking process before the final answer. Keep the SAME final answer.
\par\smallskip
Question:\\
\{q\}
\par\smallskip
Response to rewrite:\\
\{gold\}
}
\end{tcolorbox}

\begin{tcolorbox}[
    colback=gray!5!white,
    colframe=gray!75!black,
    breakable,
    title={Prompt B: No Format, on tool-use},
    label={box:prompt_b_tooluse}
]
{\ttfamily\small
Rewrite the following response in your own words. Keep the SAME final answer.
\par\smallskip
Question:\\
\{q\}
\par\smallskip
Response to rewrite:\\
\{gold\}
\par\smallskip
End with exactly this format: Action: \textless{}tool\textgreater{}\\
Action Input: \textless{}json\textgreater{}
}
\end{tcolorbox}

\begin{tcolorbox}[
    colback=gray!5!white,
    colframe=gray!75!black,
    breakable,
    title={Prompt C: Thinking, on tool-use},
    label={box:prompt_c_tooluse}
]
{\ttfamily\small
Rewrite the following response in your own words, including the thinking process. Keep the SAME final answer.
\par\smallskip
Question:\\
\{q\}
\par\smallskip
Response to rewrite:\\
\{gold\}
\par\smallskip
End with exactly this format: Action: \textless{}tool\textgreater{}\\
Action Input: \textless{}json\textgreater{}
}
\end{tcolorbox}

\section{Setup for the In-distribution Experiments}
\label{app:id-setup}

Both experiments continue fine-tuning \ttt{Qwen2.5-3B-Instruct} in \ttt{bf16} with LoRA ($r{=}16$,
$\alpha{=}32$, all linear layers) for $1$ epoch at a constant learning rate of $10^{-4}$ with AdamW,
seed $1$, and differ only in the training data. Decoding is greedy. Tab.~\ref{tab:corpora} lists every dataset Experiment 1 trains on, with its source and upload date.

\begin{table*}[t]
\centering
\footnotesize
\setlength{\tabcolsep}{5pt}
\begin{tabular}{@{}ll@{}}
\toprule
\textbf{Training Dataset (HuggingFace Path)} & \textbf{Uploaded} \\
\midrule
\addlinespace[2pt]
\rowcolor{rowtask}\multicolumn{2}{@{}l@{}}{\textbf{Pre-cutoff}: math released before the cutoff} \\
\addlinespace[1pt]
\ttt{microsoft/orca-math-word-problems-200k} & 2024-03 \\
\ttt{openai/gsm8k} (\ttt{main}) & 2022-04 \\
\ttt{AI-MO/NuminaMath-1.5} & 2025-02 \\
\ttt{EleutherAI/hendrycks\_math} (\ttt{algebra}) & 2023-09 \\
\ttt{TIGER-Lab/MathInstruct} & 2023-09 \\
\ttt{deepmind/aqua\_rat} (\ttt{raw}) & 2022-03 \\
\ttt{meta-math/MetaMathQA} & 2023-09 \\
\ttt{nvidia/OpenMathInstruct-1} & 2024-02 \\
\addlinespace[2pt]
\rowcolor{rowtask}\multicolumn{2}{@{}l@{}}{\textbf{Post-cutoff}: long-CoT reasoning released after the cutoff} \\
\addlinespace[1pt]
\ttt{zwhe99/DeepMath-103K} & 2025-04 \\
\ttt{nvidia/OpenMathReasoning} (split \ttt{cot}) & 2025-04 \\
\ttt{simplescaling/s1K-1.1} & 2025-02 \\
\ttt{GAIR/LIMO} & 2025-02 \\
\ttt{open-thoughts/OpenThoughts-114k} & 2025-01 \\
\ttt{qihoo360/Light-R1-SFTData} & 2025-03 \\
\ttt{bespokelabs/Bespoke-Stratos-17k} & 2025-01 \\
\ttt{open-r1/OpenR1-Math-220k} & 2025-02 \\
\ttt{open-thoughts/OpenThoughts2-1M} & 2025-04 \\
\ttt{a-m-team/AM-DeepSeek-R1-Distilled-1.4M} (\ttt{am\_0.5M}) & 2025-03 \\
\addlinespace[2pt]
\rowcolor{rowtask}\multicolumn{2}{@{}l@{}}{\textbf{Off-target}: code or general instruction data, also post-cutoff} \\
\addlinespace[1pt]
\ttt{OpenCoder-LLM/opc-sft-stage2} (\ttt{educational\_instruct}) & 2024-11 \\
\ttt{EricLu/SCP-116K} & 2025-01 \\
\ttt{allenai/tulu-3-sft-mixture} & 2024-11 \\
\ttt{KodCode/KodCode-V1} & 2025-02 \\
\ttt{nvidia/OpenCodeReasoning} (\ttt{split\_0}) & 2025-04 \\
\bottomrule
\end{tabular}
\caption{\textbf{The training data.} The $23$ corpora Experiment 1 trains on, each taken as
published and named by the HuggingFace path it is loaded from. Groupings are domain and upload date
against the base model's 2024-09 cutoff, and the dates are each repository's \ttt{createdAt} read
from the HuggingFace API.}
\label{tab:corpora}
\end{table*}

\section{Further Experimental Details} \label{app:exp}

\subsection{Hardware} \label{app:hw}
All experiments run on A100 nodes. For other hardware such as CPU, RAM and disk we use commonly
available ones, as these did not induce any bottlenecks.

\subsection{Datasets} \label{app:data}
All datasets are set to default values following the SDFT paper.
Tab.~\ref{tab:science-mix-split} gives the \tsc{Science Mixture} split. \tsc{SciKnowEval} supplies
only an answer key for its multiple-choice rows, so the \ttt{<reasoning>} demonstrations every arm
trains on or rewrites are model-generated: GPT-4o for \tsc{Chemistry}, released by
\citet{shenfeld2026selfdistillation}, and \ttt{gpt-5.6-luna} for the other three subjects.
\textbf{Gold} is therefore the subset of the L-3 pool a correct demonstration exists for, split
$80$/$20$ into training and test, so the evaluation prompts are filtered the same way, though every
arm is scored on the same pool.

\begin{table}[bt!]
\centering
\small
\setlength{\tabcolsep}{4pt}
\begin{tabular}{@{}lrrrrrr@{}}
\toprule
 & & \multicolumn{3}{c}{\textbf{Per-subject domain}} & \multicolumn{2}{c}{\textbf{\tsc{Science-mix}}} \\
\cmidrule(lr){3-5} \cmidrule(lr){6-7}
\textbf{Subject} & \textbf{L-3} & \textbf{Gold} & \textbf{Train} & \textbf{Test} & \textbf{Train} & \textbf{Test} \\
\midrule
\ttt{Chemistry} & 3,542 & 2,674 & 2,139 & 535 & 668 & 133 \\
\ttt{Biology} & 679 & 356 & 285 & 71 & 285 & 71 \\
\ttt{Material} & 1,283 & 969 & 775 & 194 & 668 & 133 \\
\ttt{Physics} & 1,498 & 1,114 & 891 & 223 & 668 & 133 \\
\midrule
\textbf{Total} & \textbf{7,002} & \textbf{5,113} & \textbf{4,090} & \textbf{1,023} & \textbf{2,289} & \textbf{470} \\
\bottomrule
\end{tabular}
\caption{\textbf{Train/test split of the \tsc{Science-mix} corpus.} \tsc{SciKnowEval} L-3 gold,
split $80$/$20$ at seed $0$.}
\label{tab:science-mix-split}
\end{table}

\subsection{Training Configuration} \label{app:train}
Tab.~\ref{tab:config} lists the settings. Every cell is a single seed $42$ chosen \textit{a priori}, and never changed after experimentation.

\begin{table*}[bt!]
\centering
\small
\begin{tabular}{@{}p{0.25\textwidth}p{0.71\textwidth}@{}}
\toprule
\textbf{Setting} & \textbf{Value} \\
\midrule
Adaptation & full fine-tuning of all parameters \\
Learning rate & $5\times10^{-5}$ \\
Epochs & $2$ \\
Effective batch & $16$ sequences \\
Optimizer & paged AdamW, 8-bit \\
Schedule & cosine; warmup ratio $0.1$ \\
Gradient clipping & $\lVert g \rVert \le 1.0$ \\
Seed & $42$ \\
\bottomrule
\end{tabular}
\caption{\textbf{Configuration for every cell in Tabs.~\ref{tab:main} and~\ref{tab:aligner-family}.}
Unlisted values are library defaults.}
\label{tab:config}
\end{table*}

\clearpage
\section{Qualitative Example of Aligner Rewrites} \label{app:samples}

A qualitative example is provided in Box \ref{box:sample_science_good}.

\begin{tcolorbox}[
    colback=gray!5!white,
    colframe=gray!75!black,
    breakable,
    title={Sample: Science, aligner \ttt{Qwen2.5-7B}},
    label={box:sample_science_good}
]
\textbf{Question} (the task prompt, shown to the student at train and test time)\\[2pt]
{\ttfamily\small
How do you calculate the half life of carbon 14?
\par\smallskip
A: 5.73 x 10\textasciicircum{}3 yr\\
B: 4.56 x 10\textasciicircum{}3 yr\\
C: 7.89 x 10\textasciicircum{}3 yr\\
D: 3.21 x 10\textasciicircum{}3 yr
}
\tcbline
\textbf{Gold} (the original demonstration; what plain SFT trains on)\\[2pt]
{\ttfamily\small
\textless{}reasoning\textgreater{}\\
The half-life of carbon-14 is a well-established scientific fact used in radiocarbon dating. The correct value is approximately 5730 years. This corresponds to option A: 5.73 x 10\textasciicircum{}3 yr.\\
\textless{}/reasoning\textgreater{}\\
\textless{}answer\textgreater{}\\
A\\
\textless{}/answer\textgreater{}
}
\tcbline
\textbf{Rewrite} (what \tsc{SynthSelf} trains on)\\[2pt]
{\ttfamily\small
\textless{}reasoning\textgreater{}\\
The half-life of carbon-14 is known to be about 5730 years, which matches option A: 5.73 x 10\textasciicircum{}3 yr.\\
\textless{}/reasoning\textgreater{}\\
\textless{}answer\textgreater{}\\
A\\
\textless{}/answer\textgreater{}
}
\end{tcolorbox}

\end{document}

%% file: math_commands.tex
\usepackage{amsmath,amsfonts,bm}

\def\eqref#1{equation~\ref{#1}}

\def\1{\bm{1}}

\DeclareMathAlphabet{\mathsfit}{\encodingdefault}{\sfdefault}{m}{sl}
\SetMathAlphabet{\mathsfit}{bold}{\encodingdefault}{\sfdefault}{bx}{n}

